%% file: root.tex
\documentclass[letterpaper, 10 pt, conference]{ieeeconf}  

\IEEEoverridecommandlockouts                              

\usepackage{booktabs}
\input{config}
\input{symbols}

\title{\LARGE \bf
Jointly Assigning Processes to Machines and Generating Plans for Autonomous Mobile Robots in a Smart Factory 
}

\author{Chistopher Leet$^{1}$, Aidan Sciortino$^{2}$, Sven Koenig$^{3}$
\thanks{$^{1}$University of Southern California,
        {\tt\small cjleet@usc.edu}}%
\thanks{$^{2}$University of Rochester
{\tt\small asciorti@u.rochester.edu}}%
\thanks{$^{3}$University of Southern California,
        {\tt\small skoenig@usc.edu}}%
}

\begin{document}

\maketitle
\thispagestyle{empty}
\pagestyle{empty}

\begin{abstract}
A modern smart factory runs a manufacturing procedure using a collection of programmable machines. 
Typically, materials are ferried between these machines using a team of mobile robots. 
To embed a manufacturing procedure in a smart factory, a factory operator must a) assign its processes to the smart factory's machines and b) determine how agents should carry materials between machines. 
A good embedding maximizes the smart factory's throughput; the rate at which it outputs products. 
Existing smart factory management systems solve the aforementioned problems sequentially, limiting the throughput that they can achieve. 
In this paper we introduce ACES, the Anytime Cyclic Embedding Solver, the first solver which jointly optimizes the assignment of processes to machines and the assignment of paths to agents. 
We evaluate ACES and show that it can scale to real industrial scenarios.
\end{abstract}

\begin{figure*}[t]
    \centering\includegraphics[width=\linewidth, trim={0 0 0 0},clip]{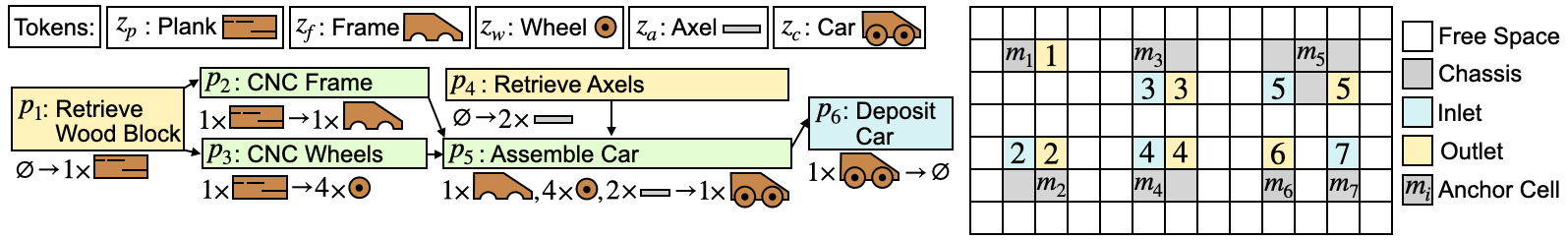}

    \vspace{-3mm}
    \caption{(left) An example manufacturing procedure. Source process are yellow; sink processes blue. (right) An example smart factory.}
    \vspace{-5mm}
    \label{fig:example_problem}
\end{figure*}

\section{INTRODUCTION}

Modern smart factories are designed to enable flexible manufacturing~\cite{enabling_flexible_manufacturing}. A flexible manufacturing system is a system which can produce a variety of different products with minimal reconfiguration~\cite{flexible_manufacturing_impact}. Flexibility can improve a manufacturer's ability to customize products, reduce the time that it takes to fulfill new orders, and lower the costs of producing a new product. Today, a wide range of industries practice flexible manufacturing, including the automotive, medical, and textile industries~\cite{flexible_manufacturing_examples}.

To permit flexible manufacturing, a smart factory needs the following two components:

\para{1) Flexible Machines.} Flexible machines are general-purpose machines such as CNC machines which can be programmed to carry out a range of manufacturing processes~\cite{trends_in_smart_manufacturing}. Their programmability makes it easy to change the process assigned to each machine when the product produced by the smart factory changes. Their programmability also makes it easy to relieve bottlenecks in a manufacturing procedure by changing the number of machines assigned to its processes.

\para{2) Flexible Transport System.} A flexible transport system makes it easy to adjust the flow of parts through a smart factory when the product produced by the smart factory changes. Typically, flexible transport systems use a team of agents to ferry parts between machines~\cite{AMRs_in_manufacturing_operations}. In a traditional smart factory, agents are autonomous mobile robots~\cite{AMRs_in_manufacturing_operations}. In a mag-lev based smart factory, such as BOSCH's ctrlX Flow$^{\text{6D}}$~\cite{BOSCH_maglev}, agents are magnetically levitating shuttles. With such a system, adjusting the flow of parts between machines is as simple as adjusting the circulation of agents through the factory.

To embed a manufacturing procedure into a  smart factory, a factory operator needs to:

\begin{enumerate}
    \item assign the processes in the manufacturing procedure to the smart factory's machines.
    \item find a transport plan which specifies how the smart factory's agents should ferry materials between machines.
\end{enumerate}

\noindent We term the problem of embedding a manufacturing procedure into a smart factory the Smart Factory Embedding Problem (SFEP). A good embedding maximizes a smart factory's throughput, the rate at which it outputs products.

To date, the SFEP is open. To our knowledge, no existing smart factory management system~\cite{sf_traffic_system,sf_dpn} jointly optimizes the assignment of processes to machines and paths to agents. Solving these problems separately limits the overall throughput that these systems achieve.

In this paper, we address this lacuna. First, we present the first formal model of the SFEP as a combinatorial optimization problem. We then solve the SFEP by introducing ACES, the Anytime Cyclic Embedding Solver. ACES models the SFEP as a mixed-integer linear program (MILP). Na\"{i}vely modeling the SFEP as a MILP does not generate a practical solution.  Multi-Agent Path Finding (MAPF), the problem of finding collision-free paths for a team of agents, is an important component of the SFEP. MILP-based solutions to MAPF often struggle to scale~\cite{MAPF_ILP_scaling}, limiting their applicability to industry. ACES addresses this problem as follows.

ACES generates cyclic transport plans. A cyclic transport plan starts and ends with its team of agents in the same state. As a result, it can be looped indefinitely. Fixing the length of a cyclic transport plan limits the number of decision variables required to generate it. Decreasing the number of decision variables in a MILP decreases its difficulty. ACES generates cyclic transport plans with incrementally longer loop lengths. The more time that ACES is given, the more loop lengths it considers and the better the throughput of its best plan.

ACES represents the parts being ferried between machines as tokens. ACES considers a variant of the SFEP where these tokens are modelled as ``agent-tokens'', abstract agents which move between machines under their own power. Two agent-tokens which represent the same part are indistinguishable. Finding a plan for a team of largely indistinguishable agent-tokens requires fewer decision variables than finding a plan for a team of regular agents. A transport plan for token agents can be converted into a transport plan for regular agents in linear time. We evaluate ACES on 6 industrial scenarios and show that it can be applied to real instances of the SFEP.

\newpage

\section{PROBLEM FORMULATION}

\subsection{Manufacturing Procedure}

\vspace{-1mm}
\para{Token.} Each raw material, part or assemblage produced or consumed during a manufacturing procedure is modelled as a token. We denote the set of tokens associated with a manufacturing procedure $\allTk{} := \{\tk{1}, \tk{2}, \ldots\}$.

\para{Process.} A process $\p{i}$ is an atomic operation in a manufacturing procedure. A process transforms a multiset of input tokens $\inTk{\p{i}}$ into a multiset of output tokens $\outTk{\p{i}}$. We denote the number of copies of token $\tk{j} \in \allTk{}$ that a process $\p{i}$ consumes and emits $\inTkType{\p{i}}{j}$ and $\outTkType{\p{i}}{j}$ respectively. 

Processes which does not consume tokens are source processes. They represent operations which retrieve raw materials. Processes which does not emit tokens are sink processes. They represent operations which export finished products or remove waste. We denote a manufacturing procedure's set of processes $\allP{} := \{\p{1}, \p{2}, \ldots\}$. Exactly one of these processes must be an output process $\outP{}$, a sink process which exports finished products. 

\para{Manufacturing Procedure.} A manufacturing procedure  $(\allTk{}, \allP{}, \outP{})$ is a set of processes $\allP{}$ which consume and emit tokens from the set $\allTk{}$. It has the output process $\outP{}$.

\para{Example.} Fig.~\ref{fig:example_problem}. (left) depicts a manufacturing procedure that makes toy cars. It has two source processes: $\p{1}$ and $\p{4}$. Its output process is $\p{6}$. Process $\p{2}$ consumes 1 plank token $\tk{p}$ and emits 1 frame token $\tk{f}$. Arrows show the flow of tokens through the manufacturing procedure.


\subsection{Smart Factory}

\vspace{-1mm}
\para{Machines.} A smart factory contains a set of machines $\allM{} := \{\m{1}, \m{2}, \ldots\}$. Each machine $\m{i} \in \allM{}$ can run a subset $\mProc{\m{i}}$ of the processes in $\allP{}$. Time  is discretized. The number of timesteps that machine $\m{i}$ takes to run a process $\p{j} \in \mProc{\m{i}}$ is denoted $\runtime{\m{i}}{\p{j}} \in \mathbb{N}$.

A machine has an input and an output buffer. When a machine starts to run a process $\p{j}$, it consumes the multiset of tokens $\inTk{\p{j}}$ from its input buffer. If its input buffer does not contain these tokens, it cannot run process $\p{j}$. When it finishes $\p{j}$, it emits the multiset of tokens $\outTk{\p{j}}$ into its output buffer. At timestep $t$, machine $\m{i}$'s input and output buffers contain the multisets of tokens $\bIn{\m{i}}{t}$ and $\bOut{\m{i}}{t}$. 

\para{Layout.} We model a factory layout as a 4-connected grid of cells. A cell is traversable if an agent can enter it and non-traversable if it contains an obstacle such as a machine chassis. The set of traversable cells is denoted $\allCl{} := \{\cl{1}, \cl{2}, \ldots\}$.  

Any machine which can run a process that consumes tokens has an input cell $\mIn{\m{i}}$. Any machine which can run a process that emits tokens has an output cell $\mOut{\m{i}}$. We denote the set of all input and output cells $\allCIn{}$ and $\allCOut{}.$ Tokens can only be placed in a machine's input buffer from its input cell and and removed from a machine's and output buffer from its output cell. A machine's input and output cells must be traversable. All input and output cells must be distinct. A machine without an input cell is a source machine. Source machines only run source processes. They represent bins of raw materials. A machine without an output cell is a sink machine. Sink machines only run sink processes. They represent output chutes and waste bins.

\begin{table}[t]
    \centering
    \begin{tabular}{l l l}
        \textbf{Machine Type}  &\textbf{Instances}   &\textbf{Supported Processes}\\
         Bin of Planks         &$\m{1}$                &$\p{1}$\\
         CNC Machine           &$\m{2}$, $\m{3}$, $\m{4}$  &$\p{2}$, $\p{3}$\\
         Assembler             &$\m{5}$                &$\p{5}$\\
         Bin of Axles          &$\m{6}$                &$\p{4}$\\
         Output Chute          &$\m{7}$                &$\p{6}$
    \end{tabular}
    \vspace{-1mm}
    \caption{The machines in the example smart factory.}
    \label{tab:example_machines}
    \vspace{-5mm}
\end{table}

\para{Example.} Fig.~\ref{fig:example_problem}. (right) depicts a smart factory. Machine $\m{2}$ has the inlet and outlet cell $(1,2)$ and $(2,2)$. Table~\ref{tab:example_machines} describes its machines and lists the processes that they can run. There are two source machines: $\m{1}$, a bin of planks, and $\m{5}$, a bin of axles, and one sink machine: $\m{6}$, an output chute.

\subsection{Agents}

Tokens are carried between machines by a team of agents $\allAg{} := \{\ag{1},  \ldots \ag{\agentNo{}}\}$. At the start of a timestep $t$, an agent $\ag{i}$ occupies a traversable cell. We denote this cell $\agCl{\ag{i}}{t}$. Each timestep, an agent must wait at its current cell or move to a traversable cell which shares an edge with its current cell.

We represent the actions that an agent can take at each cell with an undirected graph called the movement graph $\layoutG{} := (\allCl{}, \layoutE{})$. Each vertex in this graph is a traversible cell. There is an edge $(\cl{i}, \cl{j}) \in \layoutE{}$ between two cells $\cl{i}, \cl{j} \in \allCl{} \times \allCl{}$ iff an agent at cell $\cl{i}$ can be at cell $\cl{j}$ on the following timestep. Consequently, each traversible cell $\cl{i}$ is connected to itself by a loop edge and to any traversible cell it shares a side with. Two agents may not occupy the same cell or traverse the same edge in the movement graph on the same timestep.

An agent can carry a single token. We term the token that an agent $\ag{i}$ is carrying on timestep $t$ its cargo and denote it $\agTk{\ag{i}}{t}$. If agent $\ag{i}$ is not carrying a token on timestep $t$, its cargo is the null token $\tk{0}$. The set of all tokens that an agent can carry $\allTk{} \cup \{\tk{0}\}$ is denoted $\allTkWithNull{}$. The state $(\agCl{\ag{i}}{t}, \agTk{\ag{i}}{t})$ of agent $\ag{i}$ on timestep $t$ is its location and its cargo.

An agent with a (non-null) token on a machine's input cell may deposit its token into the machine's input buffer. An agent without a (non-null) token on a machine's output cell may pick up a token from the machine's output buffer. Picking up and depositing a token takes a single timestep. An agent cannot move during this timestep.

\subsection{Embedding}

An embedding describes how a manufacturing procedure  is implemented by a smart factory. An embedding is a 6-tuple $(\assignM{}, \rateM{}, \bInM{}, \bOutM{}, \agClM{}, \agTkM{})$ with the following components:


\para{Assignment Matrix.} The assignment matrix $\assignM{}$ is an $|\allM{}| \times \allP{}|$ matrix. The field $\assign{\m{i}}{\p{j}} \in \{0, 1\}$ contains a binary variable which indicates iff machine $\m{i}$ has been assigned process $\p{j}$. A machine can be assigned at most one process.

\para{Rate Matrix.} The rate matrix $\rateM{}$ is also a $|\allM{}| \times \allP{}|$ matrix. The field $\rate{\m{i}}{\p{j}}$ indicates the rate, in runs per timestep, that machine $\m{i}$ runs process $\p{j}$ at. The rate matrix allows the rate that a machine runs its process at to be decreased, synchronizing it with the rest of the factory. A machine $\m{i}$ can only run a process $\p{j}$ at a non-zero rate iff it is assigned that process.  The maximum rate that machine $\m{i}$ can run a process $\p{j} \in \mProc{\m{i}}$ at is $\runtime{\m{i}}{\p{j}}^{-1}$. A machine runs a process at less than maximum rate by idling for a short time after each run. 

\para{Transport Plan.} A transport plan $(\bInM{}, \bOutM{}, \agClM{}, \agTkM{})$ describes how tokens move through the factory. It specifies the state $(\agCl{\ag{i}}{t}, \agTk{\ag{i}}{t})$ of each agent $\ag{i} \in \allAg{}$ and the tokens in the input and output buffers $(\bIn{\m{i}}{t},\bOut{\m{i}}{t})$ of each machine $\m{i} \in \allM{}$ at each timestep $t$. 

A factory may need to run a manufacturing procedure for an indefinite amount of time. We thus need to find a cyclic transport plan~\cite{CCPP}, a transport plan which can be looped repeatedly. A cyclic transport plan has a cycle time $\ct{}$ and an agent permutation $\agP{} : \allAg{} \rightarrow \allAg{}$. It runs from timestep $t = 0$ to $t = \ct{}$. The tokens in each machine's input buffer and output buffer at $t = 0$ and $t = \ct{}$ must be the same:
\begin{align*}
    \forall\ \m{i} \in \allM{},\ \bIn{\m{i}}{0} = \bIn{\m{i}}{\ct{}} \wedge \bOut{\m{i}}{0} = \bOut{\m{i}}{\ct{}}.
\end{align*}

 An agent $\ag{i} \in \allAg{}$ must be in the same state at timestep $t = 0$ as the agent $\agP{}(\ag{i})$ at timestep $t = \ct{}$:
\begin{align*}
    &\forall\ \ag{i} \in \allAg{}, \agCl{\ag{i}}{0} = \agCl{\agP{}(\ag{i})}{T} \wedge \agTk{\ag{i}}{0} = \agTk{\agP{}(\ag{i})}{T}.
\end{align*}    

A cyclic transport plan can be looped by following the plan from $t = 0$ to $t = \ct{}$, relabelling the agents in $\allAg{}$ according to the permutation $\agP{}$, and repeating. 

\subsection{The Smart Factory Embedding Problem}

The throughput $\throughput{}(\assignM{}, \rateM{}, \bInM{}, \bOutM{}, \agClM{}, \agTkM{})$ of an embedding is the total rate at which its machines run its output process:
\begin{equation*}
     \throughput{}(\assignM{}, \rateM{}, \bInM{}, \bOutM{}, \agClM{}, \agTkM{}) := \sum_{\m{i} \in \allM{}} \frac{1}{\rate{\m{i}}{\outP{}}}.
\end{equation*}

In the Smart Factory Embedding Problem (SFEP), we are given a manufacturing procedure $(\allTk{}, \allP{}, \outP{})$ and a smart factory $(\allM{}, \layoutG{}, \allAg{})$ and asked to find a maximal throughput embedding $(\assignM{}, \rateM{}, \bInM{}, \bOutM{}, \agClM{}, \agTkM{})$.

\section{RELATED WORK}

The Multi-Agent Path Finding (MAPF) problem is the problem of moving a team of agents from their starting positions to their goal positions without a collision. MAPF is an important component of the SFEP. The MAPF problem has been solved in a number of different ways, including prioritized planning~\cite{PP}, search~\cite{CBS}, answer set programming~\cite{ASP_LMAPF_offline}, rule-based AI~\cite{push_and_rotate} and SAT solving~\cite{MAPF_SAT}.

MAPF has been studied in the context of other optimization problems. One problem that involves MAPF which is related to the SFEP is the Collective Construction Problem~\cite{CCP_Definition}. In the CCP, a team of agents is asked to construct a structure out of building blocks. These building blocks are the same size as the robots, forcing the robots to scale the structure to position the blocks. The CCP is modelled as a combinatorial optimization problem and solved for a single agent using dynamic programming in~\cite{single_agent_CCP}. This algorithm is extended to multiple agents in~\cite{dp_CCP}. Its solutions, however, achieve little parallelism. The reinforcement learning approach developed in~\cite{rl_CCP} improves parallelism. Optimal solutions to the CCP based on constraint set programming and mixed integer linear programming are proposed in~\cite{CSP_MILP_CCP}.

To our knowledge, the problem of jointly optimizing the assignment of processes to machines and the paths that agents follow to carry materials between machines in a smart factory is not well studied. We believe that we are the first to model this problem as a combinatorial optimization problem. A number of systems for coordinating mobile robots in a manufacturing plant have been proposed. For example, in~\cite{sf_traffic_system}, the authors coordinate mobile robots in a manufacturing plant using a traffic system. In~\cite{sf_dpn}, a distributed petri-net assigns tasks to mobile robots and coordinates traffic in a flexible manufacturing system. Neither of these systems, however, optimizes the assignment of processes to machines.

\section{APPROACH}

We solve the SFEP in two stages. First, we construct a simplified variant of the SFEP, which we term the Fixed cycle Length, Agent-Token SFEP (FLAT SFEP). We use this solution to construct ACES, a solution to the full SFEP.

\subsection{The Fixed Cycle Length, Agent-Token SFEP}

There are 3 differences between the FLAT and full SFEP.

\para{Difference 1. Fixed Cycle Length.} In the full SFEP, a transport plan can have any cycle length. In the FLAT SFEP,  the transport plan's cycle length is specified in the problem.

\para{Difference 2. Agent-Tokens.} In the FLAT SFEP, tokens are modeled as agents which move under their own power. Tokens in the FLAT SFEP move analogously to agents in the full SFEP. Each timestep, a token must move to an adjacent cell, remain at its current cell, or enter a machine's input buffer. Two tokens may not occupy the same cell or traverse the same edge in the movement graph on the same timestep. 

A token may only enter a machine's input buffer from its input cell.  When a token enters a buffer, it is replaced by a null token. Null tokens may not enter a buffer. Each timestep, if a machine's output cell contains a null token, the machine may replace the null token with a token from its output buffer. A token may not move on the timestep in which it is placed on the factory floor. There may only be $n$ tokens on the factory floor at any time.

\para{Difference 3. Embedding.} A FLAT SFEP embedding has a transport plan for tokens instead of agents. It is an 8-tuple $(\assignM{}, \rateM{}, \bInM{}, \bOutM{}, \atT{}, \mvT{}, \placeT{}, \removeT{})$ with 4 new components:

\para{Position Tensor.} The position tensor $\atT{}$ is a $|\allCl{}| \times (\ct{} + 1) \times |\allTkWithNull{}|$ tensor which describes the positions of the tokens on the factory floor. Let $[0..x]$ be the set of integers $\{0, 1, \ldots, x\}$. The field $\at{\cl{i}}{t}{\tk{j}} \in [0..n]$ specifies the number of copies of token $\tk{j}$ at cell $\cl{i}$ on timestep $t$. If there is more than one token at cell $\cl{i}$ on timestep $t$, a collision has occured.

\para{Movement Tensor.} Knowing the position of each token on each timestep does not tell you how the tokens move. Since copies of the same token are indistinguishable, there may be multiple ways to produce the configuration of tokens seen at timestep $t + 1$ from the configuration seen at timestep $t$. The movement tensor $\mvT{}$, a $|\layoutE{}| \times (\ct{} + 1) \times |\allTkWithNull{}|$ tensor, eliminates this ambiguity. Recall that each vertex in the movement graph has a self-loop. The field $\mv{\cl{i}}{\cl{j}}{t}{\tk{k}} \in \{0, 1\}$ indicates iff a copy of token $\tk{k}$:

\begin{itemize}
    \item moves from $\cl{i}$ to $\cl{j}$ on timestep $t$ when $\cl{i} \neq \cl{j}$
    \item waits at $\cl{i}$ on timestep $t$ when $\cl{i} = \cl{j}$
\end{itemize}


\para{Placement and Removal Tensors.} + The fields $\place{\cl{i}}{t}{\tk{j}} \in \{0, 1\}$ and $\remove{\cl{i}}{t}{\tk{j}} \in \{0,1\}$ indicate that a copy token $\tk{j}$ was removed and placed on cell $\cl{i}$ at timestep $t$.

\subsection{Why is introducing the FLAT SFEP helpful?}

The SFEP can be formulated as an Mixed Integer Linear Program (MILP). Unfortunately, this approach scales poorly. MAPF is an key part of the SFEP. Using a MILP to solve MAPF instances with dozens of agents takes a long time. As a result, solving SFEP instances with many agents using a MILP is impractically slow. The FLAT SFEP is easier to solve as a MILP than the SFEP because it involves fewer variables. A MILP formulation of the SFEP needs to contain a binary variable indicating if cell $\cl{i}$ contains agent $\ag{j}$ on timestep $t$ for every (cell, timestep, agent) combination. A SFEP instance may have dozens of agents. Its optimal embedding may contain dozens of timesteps. As a result, there may be thousands of these variables. The FLAT SFEP needs fewer of these variables since there are usually fewer tokens than agents and it has a limited number of timesteps.

\subsection{Solving the FLAT SFEP}

We solve the FLAT SFEP by formulating it as an MILP. The MILP determines the contents of the tensors $\assignM{}$, $\rateM{}$, $\atT{}$, $\mvT{}$, $\placeT{}$ and $\removeT{}$. We then select the contents of each buffer at timestep $t = 0$, fixing their contents on all other timesteps and thus the contents of the matrices $\bInM{}$ and $\bOutM{}$.

\para{Objective.} We maximize the throughput $\throughput{}$ of our embedding: 

\vspace{-3mm}
\begin{equation*}
    \max \sum_{\m{i} \in \allM{}} \rate{\m{i}}{\outP{}}
\end{equation*}
\vspace{-4mm}

\para{Constraints.} Our formulation has three types of constraints: machine configuration constraints, buffer entry and exit constraints, and token movement constraints.

\para{Machine Configuration Constraints.} These constraints specify how each machine can be configured.

\para{Constraint 1.} A machine is assigned at most 1 process.
\begin{equation*}
    \forall\ \m{i} \in \allM{}, \sum_{\p{j} \in \allP{}} \assign{\m{i}}{\p{j}} \leq 1.
\end{equation*}

\para{Constraint 2.} A machine must able to run its process.
\begin{equation*}
    \forall\ \m{i} \in \allM{},\ \forall\ \p{j} \in \allP{} \setminus \mProc{\m{i}},\  \assign{\m{i}}{\p{j}} =0.
\end{equation*}

\para{Constraint 3.} Machine $\m{i}$ can only run a process $\p{j} \in \mProc{\m{i}}$ once every $\runtime{\m{i}}{\p{j}}$ timesteps.
\begin{equation*}
    \forall \m{i} \in \allM{},\ \forall \p{j} \in \mProc{\m{i}},  \rate{\m{i}}{\p{j}} \leq \runtime{\m{i}}{\p{j}}^{-1}
\end{equation*}

\para{Constraint 4.} Machine $\m{i}$ can only run process $\p{j}$ at a non-zero rate if machine $\m{i}$ is assigned process $\p{j}$. 
\begin{equation*}
    \forall\ \m{i} \in \allM{},\ \forall\ \p{j} \in \allP{},\ \rate{\m{i}}{\p{j}} - \assign{\m{i}}{\p{j}} \leq 0.
\end{equation*}



\para{Token Movement Constraints.} These constraints specify how tokens move through the factory.

\para{Constraint 5.} Over the course of a transport plan, the number of copies of each non-null token $\tk{j} \in \allTk{}$ that machine $\m{i}$ consumes and that enter its input buffer must be the same.
\begin{align*}
    &\forall\ \m{i} \in \allM{},\ \forall\ \tk{j} \in \allTk{},\\%
    &\sum_{t \in [0..\ct{}]} \remove{\mIn{\m{i}}}{t}{\tk{j}}
    = \sum_{\mathclap{\p{k} \in \allP{}}} \rate{\m{i}}{\p{j}} \cdot \inTkType{\p{k}}{j} \cdot \ct{}.%
\end{align*}

\para{Constraint 6.} Over the course of a transport plan, the number of copies of each non-null token $\tk{j} \in \allTk{}$ that machine $\m{i}$ emits and that
exit its output buffer must be the same.
\begin{align*}
    &\forall\ \m{i} \in \allM{},\ \forall\ \tk{j} \in \allTk{},\\%
    &\sum_{t = [0..\ct{}]} \place{\mOut{\m{i}}}{t}{\tk{j}}
    = \sum_{\mathclap{\p{k} \in \allP{}}} \rate{\m{i}}{\p{j}} \cdot \outTkType{\p{k}}{j} \cdot \ct{}.%
\end{align*}

\para{Constraint 7.} Each timestep $t$, a token $\tk{j} \in \allTkWithNull{}$ must wait at its cell, move to an adjacent cell, or be removed from the factory floor. Recall that each vertex in the movement graph has a self-loop, and that the field $\mv{\cl{i}}{\cl{i}}{t}{\tk{j}}$ indicates if token $\tk{j}$ waits at cell $\cl{i}$ on timestep $t$.
\begin{align*}
\forall\ \cl{i}, t&, \tk{j} \in C \times [0..\ct{}] \times \allTkWithNull{},\ \at{\cl{i}}{t}{\tk{j}} = \\
&\sum_{\mathclap{(\cl{i},\cl{k}) \in \layoutE{}}} \mv{\cl{i}}{\cl{k}}{t}{\tk{j}} + 
\begin{cases}
\remove{\cl{i}}{t}{\tk{j}} &\cl{i} \in \allCIO{}\\
0               &\text{otherwise}
\end{cases}.
\end{align*}

\para{Constraint 8.} The number of copies of token $\tk{j} \in \allTkWithNull{}$ at cell $\cl{i}$ on timestep $t + 1$ is equal to the number of copies of $\tk{j}$ that wait at, move to, or are placed on cell $\cl{i}$ on timestep $t$.
\begin{align*}
\forall\ \cl{i}, t&, \tk{j} \in C \times [0..\ct{}] \times \allTkWithNull{},\ \at{\cl{i}}{t + 1}{\tk{j}} = \\
&\sum_{\mathclap{(\cl{k}, \cl{i}) \in \layoutE{}}} \mv{\cl{k}}{\cl{i}}{t}{\tk{j}} + 
\begin{cases}
\place{\cl{i}}{t}{\tk{j}} &\cl{i} \in \allCIO{}\\
0               &\text{otherwise}
\end{cases}.
\end{align*}

\newpage
\para{Constraint 9.} A cell may not contain two tokens at once.

\vspace{-5mm}
\begin{align*}
\forall\ \cl{i}, t \in \allCl{} \times [0..\ct{}],\ \sum_{\tk{k} \in \allTkWithNull{}} \at{\cl{i}}{t}{\tk{j}} \leq 1.
\end{align*}
\vspace{-2mm}

\noindent Together, constraints 7, 8 and 9 imply that at most one token can be placed on or removed from a cell on any timestep.

\para{Constraint 10.} If a non-null token $\tk{} \in \allTk{}$ is removed from an input cell, it must be replaced by a null token $\tk{0}$.

\vspace{-4mm}
\begin{equation*}
\forall\ \cl{i}, t \in \allCIn{} \times [0..\ct{}],\ \sum_{\tk{k} \in \allTk{}} \remove{\cl{i}}{t}{\tk{j}} = \place{\cl{i}}{t}{\tk{0}}.
\end{equation*}
\vspace{-3mm}

\para{Constraint 11.} If a non-null token $\tk{} \in \allTk{}$ is placed on an output cell, it must replace a null token $\tk{0}$.

\vspace{-4mm}
\begin{equation*}
\forall\ \cl{i}, t \in \allCOut{} \times [0..\ct{}],\ \sum_{\tk{k} \in \allTk{}} \place{\cl{i}}{t}{\tk{j}} = \remove{\cl{i}}{t}{\tk{0}}.
\end{equation*}
\vspace{-3mm}

\para{Constraint 12.} A non-null token $\tk{k} \in \allTk{}$ cannot be placed on an input cell or removed from an output cell.

\vspace{-4mm}
\begin{align*}
&\forall\ \cl{i}, t, \tk{j} \in \allCIn{} \times [0..\ct{}] \times \allTk{},\ \place{\cl{i}}{t}{\tk{j}} = 0.\\
&\forall\ \cl{i}, t, \tk{j} \in \allCOut{} \times [0..\ct{}] \times \allTk{},\  \remove{\cl{i}}{t}{\tk{j}} = 0.\\
\end{align*}
\vspace{-11mm}

\para{Constraint 13.} A null token $\tk{0}$ cannot be placed on an output cell or removed from an input cell.

\vspace{-5mm}
\begin{align*}
&\forall\ \cl{i}, t \in \allCOut{} \times [0..\ct{}],\ \place{\cl{i}}{t}{\tk{0}} = 0.\\ 
&\forall\ \cl{i}, t \in \allCIn{} \times [0..\ct{}],\ \remove{\cl{i}}{t}{\tk{0}} = 0.\\
\end{align*}
\vspace{-11mm}

\para{Constraint 14.} Two tokens may not traverse the same edge in the movement graph on the same timestep.

\vspace{-5mm}
\begin{align*}
\forall\ (\cl{i}, \cl{j}),\ &t \in \layoutE{} \times [0..\ct{}],\\ &\sum_{\tk{k} \in \allTkWithNull{}} \mv{\cl{i}}{\cl{j}}{t}{\tk{k}} + \mv{\cl{j}}{\cl{i}}{t}{\tk{k}} \leq 1.
\end{align*}
\vspace{-2mm}

\para{Constraint 15.} At most $\agentNo{}$ tokens may be on the factory floor.

\vspace{-5mm}
\begin{align*}
\sum_{\cl{i} \in \allCl{}} \sum_{\tk{j} \in \allTkWithNull{}} \at{\cl{i}}{t}{\tk{j}} \leq \agentNo{}
\end{align*}
\vspace{-2mm}

\begin{algorithm}[t]
\small
\caption{\SFEPSolver{}($\allTk{}, \allP{}, \outP{}, \allM{}, \layoutG{}, \timer{}$)}
\label{alg:SFEPAlg}
\begin{algorithmic}[1]
\STATE $\bestEmbedding{} \gets \nullSym{}$
\COMMENT{Best embedding constructed}
\STATE $\bestThroughput{} \gets 0$ 
\COMMENT{Throughput of best embedding}
\STATE $\ct{} \gets 1$ 
\COMMENT{Current cycle time}

\WHILE{$\timer{}$ has not run out of time}
\label{ln:checkTimer}
  \STATE $\embedding{}, \throughput{} \gets  \FLATSFEPSolver{}(\allTk{}, \allP{}, \outP{}, \allM{}, \layoutG{}, \ct{}, \timer{})$
  \label{ln:solveFLATSFEP}
  \IF{$\throughput{} \neq \nullSym{} \wedge \throughput{} > \bestThroughput{}$}
    \STATE $\bestEmbedding{}, \bestThroughput{} \gets \embedding{},\throughput{}$
  \ENDIF
  \STATE $\ct{} \gets \ct{} + 1$
  \label{ln:incrementCycleTime}
\ENDWHILE
\RETURN $\convertToSFEP(\bestEmbedding{})$
\label{ln:convertToSFEP}
\end{algorithmic}
\end{algorithm}

\para{Selecting the Initial Contents of a Buffer.} We initialize each machine's input buffer with the multiset of tokens that it consumes during the transport plan. As a result, even if replacement tokens arrive late, a machine will always have enough tokens to run its process. We initialize each machine's output buffer with the multiset of tokens that it emits during the transport plan for similar reasons.

\subsection{Solving the SFEP}

We use our solution to the FLAT SFEP to construct ACES. ACES is given in Algorithm~\ref{alg:SFEPAlg}. Let $\timer{}$ be a timer which triggers an interrupt after a specified amount of time. Let $\FLATSFEPSolver{}(\allTk{}, \allP{}, \outP{}, \allM{}, \layoutG{}, \ct{}, \timer{})$ (Line~\ref{ln:solveFLATSFEP}) be a implementation of our solution to the FLAT SFEP which returns an embedding and its throughput $(\embedding{}, \throughput{})$ if  successful and the tuple $(\nullSym{}, \nullSym{})$ if interrupted by the timer. Finally, let $\convertToSFEP(\bestEmbedding{})$ be a function which converts a FLAT SFEP embedding into a full SFEP embedding with the same throughput (Line~\ref{ln:convertToSFEP}). ACES's main loop (Lines~\ref{ln:checkTimer}-\ref{ln:incrementCycleTime}) solves the FLAT SFEP problem for incrementally higher values of $\ct{}$ until time runs out. ACES then converts the best embedding found into a full SFEP embedding and returns it (Line~\ref{ln:convertToSFEP}).

\begin{algorithm}[t]
\small
\caption{\convertToSFEP{}$(\assignM{}, \rateM{}, \bInM{}, \bOutM{}, \atT{}, \mvT{}, \placeT{}, \removeT{})$}
\label{alg:toSFEPEmb}
\begin{algorithmic}[1]
\STATE $\agClM{} \gets$ an empty $n \times (T+1)$ matrix
\COMMENT{Stores agent positions}
\STATE $\agTkM{} \gets$ an empty $n \times (T+1)$ matrix
\COMMENT{Stores agent cargo}
\STATE $i \gets 0$ 
\COMMENT{Agent index}
\FOR{$(\cl{j}, \tk{k}) \in \allCl{} \times \allTk{}$}
  \IF{$\at{\cl{j}}{0}{\tk{k}} = 1$}
  \label{ln:find_agent_starts}
  
    \STATE $\agCl{\ag{i}}{0} \gets \cl{j}$
    \STATE $\agTk{\ag{i}}{0} \gets \tk{k}$
    \label{ln:initialize_agents}

    \FOR{$t$ from $0$ to $\ct{}$}

      \FOR{$(\agCl{\ag{i}}{t}, \cl{l}) \in \layoutE{}$}
      \label{ln:move_check}
        \IF{$\mv{\agCl{\ag{i}}{t}}{\cl{l}}{t}{\agTk{\ag{i}}{t}} = 1$}
            \STATE $\agCl{\ag{i}}{t + 1} \gets \cl{l}$
            \STATE $\agTk{\ag{i}}{t + 1} \gets \agTk{\ag{i}}{t}$
            \label{ln:move_action}
        \ENDIF
      \ENDFOR

      \IF{$\remove{\agCl{\ag{i}}{t}}{t}{\agTk{\ag{i}}{t}}$}
      \label{ln:replace_check}
        \FOR{$\tk{m} \in \allTkWithNull{}$}
          \IF{$\place{\agCl{\ag{i}}{t}}{t}{\tk{m}}$}
            \STATE $\agCl{\ag{i}}{t + 1} \gets \agCl{\ag{i}}{t}$
            \STATE $\agTk{\ag{i}}{t + 1} \gets \tk{m}$
            \label{ln:replace_action}
          \ENDIF
        \ENDFOR
      \ENDIF

      \STATE $i \gets i + 1$
      
    \ENDFOR 
    
  \ENDIF 
\ENDFOR 

\RETURN $(\assignM{}, \rateM{}, \bInM{}, \bOutM{}, \agClM{}, \agTkM{})$
\end{algorithmic}
\end{algorithm}

\para{Converting a FLAT SFEP Embedding to a SFEP Embedding.} To convert a FLAT SFEP embedding into a SFEP embedding, we need to convert its transport plan for agent-tokens into a transport plan for regular agents. Our conversion algorithm is shown in Algorithm~\ref{alg:toSFEPEmb}. If cell $\cl{j}$ contains a copy of token $\tk{k}$ at $t = 0$ in the FLAT SFEP embedding, we position an agent $\ag{i}$ carrying a copy of token $\tk{k}$ on cell $\cl{j}$ at $t = 0$  (Lines~\ref{ln:find_agent_starts}-\ref{ln:initialize_agents}). Agent $\ag{i}$ follows this token as it moves through the factory (Lines~\ref{ln:move_check}-\ref{ln:move_action}). If this token is replaced with a copy of a new token $\tk{m}$, agent $\ag{i}$ replaces its cargo with a copy of token $\tk{m}$ by picking up or depositing a token and then starts following this new token (Lines~\ref{ln:replace_check}-\ref{ln:replace_action}).

\begin{figure*}[ht]
    \centering
    \includegraphics[width=\linewidth]{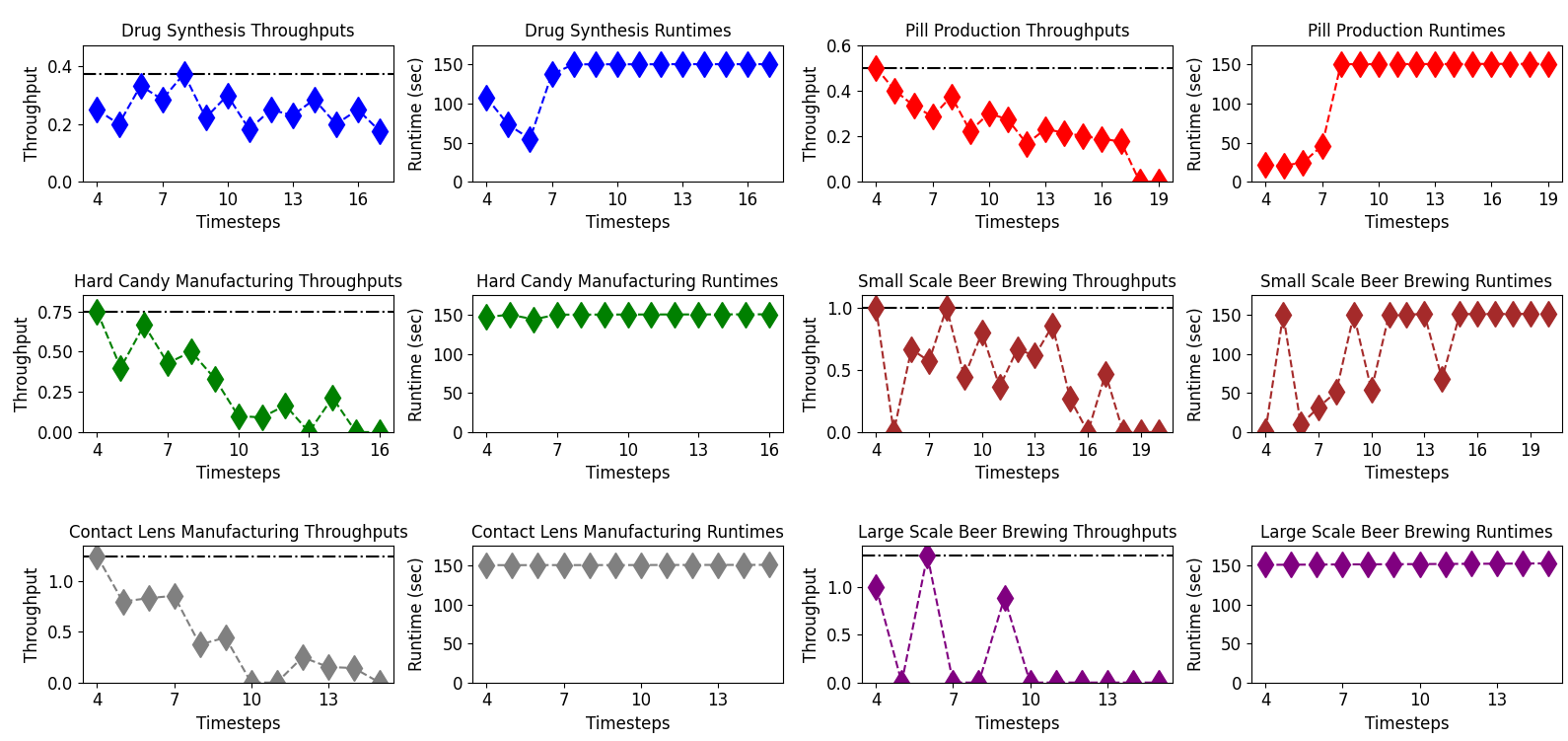}

    \vspace{-2mm}
    \caption{The FLAT SFEP solver's throughput and runtime on the Benchmark Scenarios.}
    \vspace{-4mm}
    \label{fig:aces-benchmark}
\end{figure*}

\subsection{Optimizing ACES}

ACES is optimized in two ways.

\para{Ignoring Short Cycle Lengths.} A cyclic transport plan must move each agent $\ag{i} \in \allAg{}$ from its start position to agent $\agP{}(\ag{i})$'s starting position. Consequently, agent $\ag{i}$ and $\agP{}(\ag{i})$'s starting positions can be at most $\ct{}$ cells apart. When $\ct{}$ is small, a transport plan must use relays of agents to cross large distances. These relays require an impractical number of agents to construct. Consequently, we only generate transport plans with a cycle length greater than $4$.


\para{Limiting the FLAT SFEP's Runtime.} A FLAT SFEP instance's cycle length can make it difficult to solve. ACES often gets stuck on these hard instances. Limiting the FLAT SFEP's runtime by asking it to return the best solution found in a given time frame increases the range of cycle length that ACES examines, which can improve its solution quality.

\section{EVALUATIONS}

\para{Implementation.} We implement ACES in Python 3.11~\cite{python_3_11}. We represent the layout graph with the NetworkX~\cite{networkx} library, and solve the FLAT SFEP as an ILP using Gurobi~\cite{gurobi}.

\begin{table}[t]
\begin{tabular}{llll}
\toprule
Scenario Name                                   & Processes & Machines & Max Agents \\\midrule
Drug Synthesis~\cite{API_generation}            & 8         & 16       & 100\\
Pill Production~\cite{pill_production}          & 8         & 24       & 100\\
Hard Candy Man.~\cite{hard_candy_production}   & 8         & 20       & 100\\
Small-Scale Brewing~\cite{beer}                 & 12        & 16       & 100\\
Contract Lens Man.~\cite{contact_lens_production}      & 6         & 24       & 100\\
Large-Scale Brewing~\cite{beer}                 & 12        & 23       & 100\\\bottomrule
\end{tabular}
\caption{Benchmark Scenario Details}
\vspace{-4mm}
\label{tbl:benchmark_scenarios}
\end{table}

\para{Methodology.} ACES is evaluated on 6 scenarios taken from the pharmaceutical and food manufacturing industries. These industries were chosen since they often require manufacturers to produce many slightly different variations of the same product. A hard candy manufacturer, for example, often wants to produce range of candy with different flavors. As a result, these industries benefit heavily from flexible manufacturing. Table~\ref{tbl:benchmark_scenarios} lists the number of processes and machines and maximum number of agents allowed in each scenario.
ACES was given a time limit of 30 minutes in each experiment. This time limit is realistic since embeddings are computed offline. Its FLAT SFEP solver was set to time out after 2.5 minutes. 


\para{Experimental Hardware.} Each evaluation was performed on a 3.2 GHz, 8 Core AMD Ryzen 5800H CPU with 14 GB of RAM running Ubuntu 20.04.6 LTS.

\para{Results.} The 1st and 3rd column of graphs show how the cycle length of a FLAT SFEP instance affects the throughput that the FLAT SFEP solver achieves in each scenario. A throughput of 0.0 indicates that the FLAT SFEP solver could not solve that instance. The throughput of the FLAT SFEP solver's overall best embedding in each scenario is indicated with a black line. There is little correlation between the cycle length of a FLAT SFEP instance and the throughput of its best solution. Our FLAT SFEP solver's solution quality decreases as the cycle length of a FLAT SFEP instance increases, however, because it begins to time out before finding an optimal solution. The FLAT SFEP solver never produced a solution that used all 100 agents, suggesting that its throughput was limited by its machines.

The 2nd and 4th column of graphs show how the cycle length of a FLAT SFEP instance affects the runtime of the FLAT SFEP solver. Note that the FLAT SFEP solver timed out after 150 seconds. Increasing the cycle length of an FLAT SFEP instance usually increases the FLAT SFEP solver's runtime. Interestingly, however, this is not always true. Certain cycle lengths make some FLAT SFEP instances very easy to solve.

\section{CONCLUSION}
 In this paper, we introduced the SFEP, formulated it as a combinatorial optimization problem, and addressed it with ACES, the Anytime Cyclic Embedding Solver. We see two directions for future work. First, we plan to address larger SFEP instances. Large automotive plants may operate hundreds of machines. ACES cannot scale to SFEP instances of that size. Second, we plan to allow machines to multi-task between processes. Multi-tasking is increasingly common in modern smart factories, but we have neglected it in this paper.


\bibliographystyle{IEEEtran}
\bibliography{IEEEabrv,references}

\end{document}

%% file: config.tex
\usepackage{algorithm}
\usepackage[noend]{algorithmic}
\usepackage{amssymb}
\usepackage{mathtools}

%% file: symbols.tex
\newcommand{\para}[1]{\smallskip \noindent \textit{#1}}

\newcommand{\allTk}[0]{Z}
\newcommand{\tk}[1]{z_{#1}}
\newcommand{\allP}[0]{P}
\newcommand{\p}[1]{p_{#1}}
\newcommand{\inTk}[1]{in(#1)}
\newcommand{\inTkType}[2]{in_{#2}(#1)}
\newcommand{\outTk}[1]{out(#1)}
\newcommand{\outTkType}[2]{out_{#2}(#1)}
\newcommand{\outP}[0]{\textsc{out}(P)}

\newcommand{\allM}[0]{M}
\newcommand{\m}[1]{m_{#1}}
\newcommand{\mProc}[1]{\mathcal{P}(#1)}
\newcommand{\runtime}[2]{\textsc{runtime}(#1, #2)}
\newcommand{\bInM}[0]{\boldsymbol{\mathcal{I}}}
\newcommand{\bIn}[2]{\mathcal{I}(#1, #2)}
\newcommand{\bOutM}[0]{\boldsymbol{\mathcal{O}}}
\newcommand{\bOut}[2]{\mathcal{O}(#1, #2)}
\newcommand{\allCl}[0]{C}
\newcommand{\cl}[1]{c_{#1}}
\newcommand{\agentNo}[0]{n}
\newcommand{\layoutG}[0]{G}
\newcommand{\layoutE}[0]{E}

\newcommand{\mIn}[1]{C_{in}(#1)}

\newcommand{\mOut}[1]{C_{out}(#1)}
\newcommand{\ag}[1]{a_{#1}}
\newcommand{\allAg}[0]{A}
\newcommand{\agClM}[0]{\boldsymbol{\pi}}
\newcommand{\agCl}[2]{\pi(#1, #2)}
\newcommand{\agTkM}[0]{\boldsymbol{\sigma}}
\newcommand{\agTk}[2]{\sigma(#1, #2)}

\newcommand{\allTkWithNull}[0]{Z_0}

\newcommand{\assignM}[0]{\boldsymbol{\mathcal{A}}}
\newcommand{\assign}[2]{\mathcal{A}(#1, #2)}
\newcommand{\rateM}[0]{\boldsymbol{\mathcal{R}}}
\newcommand{\rate}[2]{\mathcal{R}(#1, #2)}
\newcommand{\ct}[0]{T_c}
\newcommand{\agP}[0]{\Omega}

\newcommand{\throughput}[0]{\theta}

\newcommand{\atT}[0]{\boldsymbol{at}}
\newcommand{\at}[3]{at(#1, #2, #3)}
\newcommand{\mvT}[0]{\boldsymbol{mv}}
\newcommand{\mv}[4]{mv(#1, #2, #3, #4)}

\newcommand{\allCIn}[0]{\allCl{}_{in}}
\newcommand{\allCOut}[0]{\allCl{}_{out}}
\newcommand{\allCIO}[0]{C_{io}}
\newcommand{\placeT}[0]{\boldsymbol{pl}}
\newcommand{\place}[3]{pl(#1, #2, #3)}
\newcommand{\removeT}[0]{\boldsymbol{rm}}
\newcommand{\remove}[3]{rm(#1, #2, #3)}

\newcommand{\SFEPSolver}[0]{\textsc{ACES}}

\newcommand{\embedding}[0]{emb}
\newcommand{\bestEmbedding}[0]{\embedding{}^\ast}
\newcommand{\nullSym}[0]{\textsc{NULL}}
\newcommand{\bestThroughput}[0]{\throughput{}^\ast}
\newcommand{\timer}[0]{timer}

\newcommand{\FLATSFEPSolver}[0]{\textsc{FLATSFEP}}
\newcommand{\convertToSFEP}[0]{\textsc{ToSFEPEmb}}